\documentclass[letterpaper, 10 pt, conference]{ieeeconf}  

\IEEEoverridecommandlockouts                              

\usepackage{epsfig} 
\usepackage{mathptmx} 
\usepackage{times} 
\usepackage{amsmath} 
\usepackage{amssymb}  
\usepackage{graphicx}
\usepackage{multirow}

\usepackage[switch]{lineno}

\title{\LARGE \bf
Mixed Control for Whole-Body Compliance of a Humanoid Robot
}

\author{Xiaozhu Ju$^{1}$, Jiajun Wang$^{1}$, Gang Han$^{1}$ and Mingguo Zhao$^{2}$
\thanks{$^{1}$Xiaozhu Ju, Jiajun Wang, and Gang Han are with Beijing Research Institute of UBTECH Robotics, Beijing, China.
        {\tt\small \{xiaozhu.ju, jiajun.wang, gang.han\}@ubtrobot.com}}%
\thanks{$^{2}$Mingguo Zhao is with the Department of Automation, Tsinghua University, Beijing, China.
        {\tt\small mgzhao@mail.tsinghua.edu.cn}}%
}

\begin{document}

\maketitle
\thispagestyle{empty}
\pagestyle{empty}

\begin{abstract}
The hierarchical quadratic programming (HQP) is commonly applied to consider strict hierarchies of multi-tasks and robot's physical inequality constraints during whole-body compliance. However, for the one-step HQP, the solution can oscillate when it is close to the boundary of constraints. It is because the abrupt hit of the bounds gives rise to unrealisable jerks and even infeasible solutions. This paper proposes the mixed control, which blends the single-axis model predictive control (MPC) and proportional derivate (PD) control for the whole-body compliance to overcome these deficiencies. The MPC predicts the distances between the bounds and the control target of the critical tasks, and it provides smooth and feasible solutions by prediction and optimisation in advance. However, applying MPC will inevitably increase the computation time. Therefore, to achieve a 500 Hz servo rate, the PD controllers still regulate other tasks to save computation resources. Also, we use a more efficient null space projection (NSP) whole-body controller instead of the HQP and distribute the single-axis MPCs into four CPU cores for parallel computation. Finally, we validate the desired capabilities of the proposed strategy via Simulations and the experiment on the humanoid robot Walker X.

\end{abstract}

\section{Introduction}

Close interaction between humanoid and human gives rise to safety concerns. Hence, whole-body compliance is an essential function for the humanoid robot with a service purpose. Previous researches achieved compliant behaviour by applying admittance control \cite{c1}, \cite{c2}, \cite{c3}, \cite{c4}. In our case, we measure the external force via 6-axis force/torque sensors equipped at the wrists of the humanoid robot, and the forces are transformed to position references to fulfil the compliant behaviour. But, when the robot cooperates with people without knowledge of robotics, its constraints can be violated. Thus, the HQP is employed since this method can explicitly consider the inequality constraints. Moreover, HQP has strict prioritised task hierarchies, which ensures the compliance task does not interfere with other tasks \cite{c5}, \cite{c6}.

However, the HQP is formulated only according to the states of the immediate one servo-step, and a regular motion inequality constraint update is via a numerical integration approach \cite{c7}, \cite{c8}, \cite{c9}. For these reasons, Figure 1 shows that the one-step HQP solution can possess significant jerks or even be infeasible when a considerable external force acts on the robot. Therefore, an intuitive solution for this problem is to provide a relatively larger time interval than the sampling time \cite{c9}.

This paper utilises MPC to predict the distances between the bounds and the control target. The MPC optimises the control action several servo steps in advance to avoid the abrupt hit of bounds. Furthermore, this MPC uses an augmented state-space model of a single-axis inertia system to restrict and optimise the jerk by considering the jerk as an optimisation variable. Thus, this strategy can provide smooth and feasible solutions by prediction and optimisation in advance.

MPC is an optimisation problem that uses future information to choose the best control actions and improve controller performance. Using future information is an essential concept for controlling a humanoid robot. For instance, Kajita et al. used preview control \cite{c10}, and Wieber used receding horizon control \cite{c11} to track the zero-moment-point (ZMP) reference. Koenemann et al. applied MPC for the whole-body multi-contact and reaching problem of HRP-2, but the MPC computation is offline the robot \cite{c12}. Guo et al. formulated an offline MPC problem for the dynamic bipedal locomotion, and it was running online by approximation and interpolation \cite{c13}. The offline applications are due to the MPC's substantial computation load, which is challenging for real-time computation. In many real-time MPC applications, e.g., \cite{c14}, \cite{c15}, \cite{c16} and \cite{c17}, MPC modules run at a lower rate than the whole-body control rate because of the same reason. For instance, Yuan-Li tried to restrict the jerk with minimum-jerk MPC at a 100 Hz control rate \cite{c18}. Different from the above applications of MPC, this work hybrid uses the MPC and the PD controllers, terms as mixed control. The mixed controllers run at the same 500 Hz servo rate as the whole-body control loop. In this work, the tasks with inequality constraints use a group of single-axis model predictive controllers, i.e., ZMP, elbows and knees constraints. PD controllers regulate other tasks to save computation resources.

\begin{figure}[t]
\centering
\includegraphics[width=8.65cm]{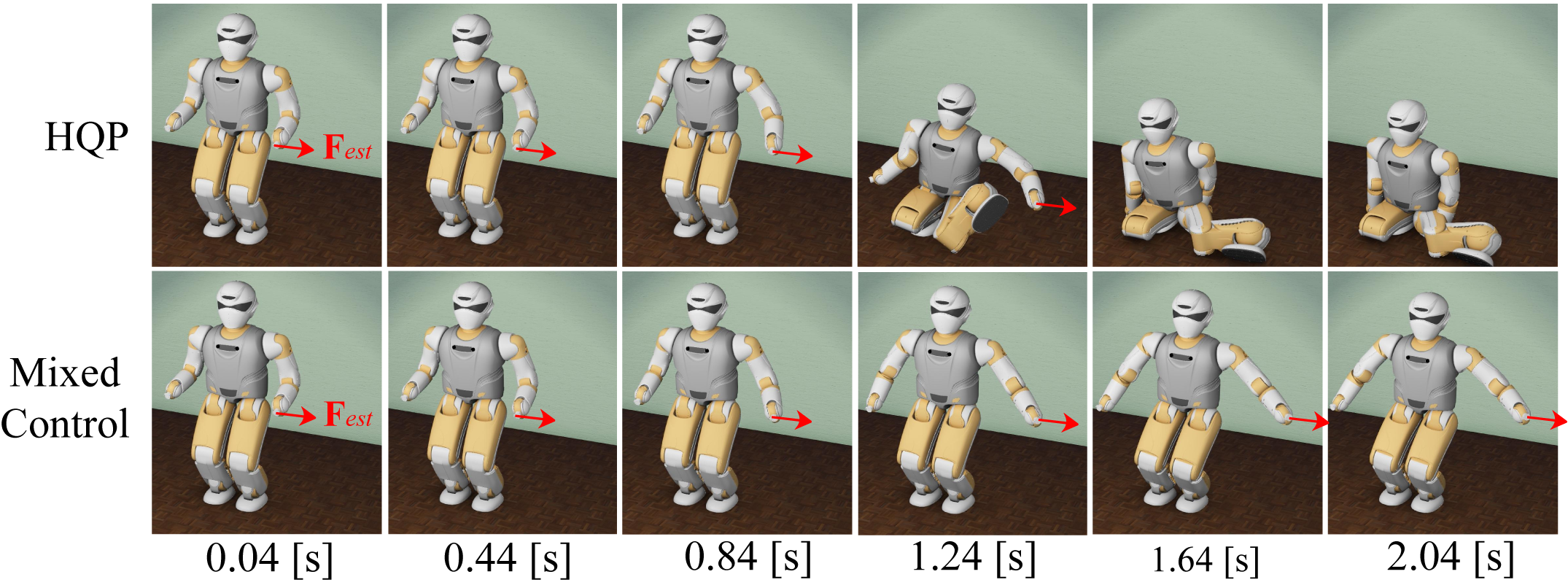}
\caption{Simulation of the whole-body compliance}
\label{figure:1}       
\end{figure}

Since the MPC can already consider inequality constraints, and the equality quadratic programming is identical to the NSP based method \cite{c19}, we use NSP instead of HQP to save the computation resources further. In addition, the NSP method decouples the tasks into different hierarchies and directions, so we can apply single-axis MPCs to control tasks and make them multi-thread for parallel computation.  We use the dynamically consistent NSP method for the whole-body control of our humanoid robot since this method is theoretically coherent and clear. Moreover, it provides a strict task hierarchy, and its computation time is promising \cite{c20}, \cite{c21}, \cite{c22}. Various humanoid robots realised remarkable achievements by utilising this method, e.g., ASIMO \cite{c23} \cite{c24}, Valkyrie \cite{c25}, DLR-Toro \cite{c26} \cite{c27} and REEM-C \cite{c4}.

In summary, the main contribution is the proposed mixed control which combines the MPC and PD controllers for whole-body compliance. The mixed control adequately resolves the deficiencies of HQP. Inevitably, the application of MPC increases the computation load. To achieve 500 Hz real-time control, we use more efficient NSP instead of HQP and distribute the single-axis MPCs into four CPU cores for parallel computation. The rest organisation of the paper is as follows. Section II describes the mathematical formulation of the mixed control, including the single-axis MPC. Section III describes the implementation of the mixed control to the NSP to obtain the joint torques for execution. This section also shows the formulation of inequality constraints in both joint space and task space. In section IV, the simulations and the experiment of the whole-body compliance demonstrate the feasibility of the proposed strategy. Finally, Section V is the conclusion and our future work.

\section{The Mixed Control}

\subsection{The Control Structure}

Figure 2 shows the control structure for the whole-body compliance. The 6-axis force/torque sensors at the wrists measure the external forces $\mathbf{F}_{ext}$. The admittance module transforms the forces $\mathbf{F}_{ext}$ to the position and orientation references $\mathbf{x}_{ref}$. Dynamic mappings of the forces are also established to the movements of the whole-body Centre of Mass (CoM) and the upper-body pose. During this process, significant accelerations generated by the compliant behaviour can break the ZMP. Moreover, we need to restrict the elbows and knees to avoid undesired configurations, such as the upper elbow configuration. Hence, we prescribe the ZMP, elbows and knees as critical tasks in this application and introduce mixed control to satisfy the inequality constraints better. The mixed control is of the form:

\begin{equation*}
\begin{aligned}
    \ddot{\mathbf{x}}_{task} = 
    \begin{cases}
    \mathbf{U} \quad &critical \  tasks \\
    \mathbf{K}_p \left( \mathbf{x}_{ref} - \mathbf{x} \right) \\
    \quad + \mathbf{K}_d \left( \dot{\mathbf{x}}_{ref} - \dot{\mathbf{x}} \right) \quad &normal \  tasks,
    \end{cases}
\end{aligned}\eqno{(1)}
\end{equation*}
where $\ddot{\mathbf{x}}_{task}$ denotes control actions in the operational space, the set $\mathbf{U}$ is the control actions generated via single-axis MPC controllers for critical tasks with inequality constraints. The diagonal matrices $\mathbf{K}_p$ and $\mathbf{K}_d$ are closed-loop gains of PD controllers for normal tasks.

\begin{figure}[tb]
\centering
\includegraphics[width=8.5cm]{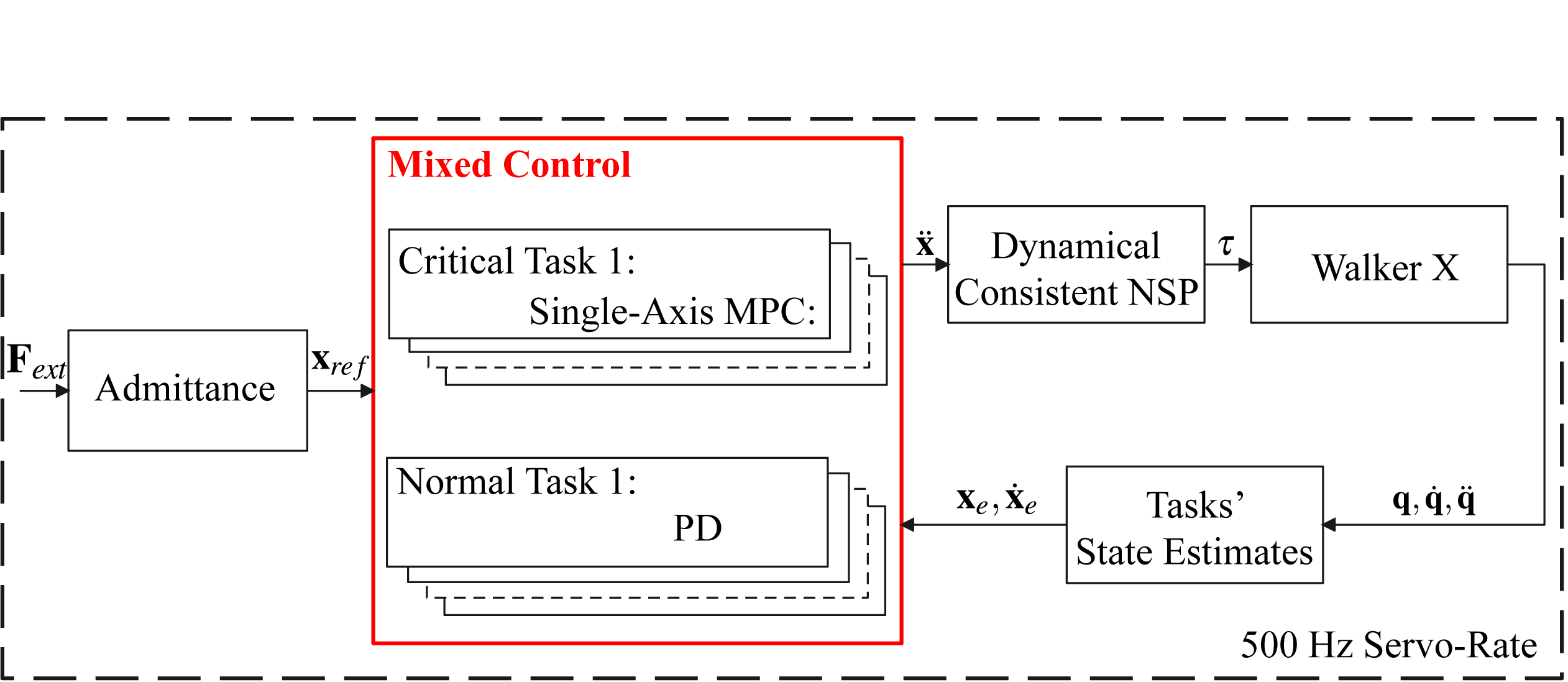}
\caption{Control structure for the whole-body compliance}
\label{figure:2}       
\end{figure}

\subsection{The Single-Axis MPC}

As previously mentioned, we utilise the single-axis MPC to resolve the deficiencies of the HQP. In this section, we aim to explain the formulation of the single-axis MPC. First, consider the discrete state space of the single-axis inertia model as:

\begin{equation*}
\begin{aligned}
    \begin{bmatrix}
    x \\
    \dot{x}
    \end{bmatrix}^{n+1} &=
    \underbrace{
    \begin{bmatrix}
    1 & \Delta t \\
    0 & 1
    \end{bmatrix}}_{\mathbf{A}_d}
    \begin{bmatrix}
    x \\
    \dot{x}
    \end{bmatrix}^n +
    \underbrace{
    \begin{bmatrix}
    \frac{\Delta t^2}{2\bar{m}} \\
    \frac{\Delta t}{\bar{m}}
    \end{bmatrix}}_{\mathbf{B}_d} U^n, \\
    y^n &= 
    \underbrace{
    \begin{bmatrix}
    1, 0
    \end{bmatrix}}_{\mathbf{C}_d}
    \begin{bmatrix}
    x \\
    \dot{x}
    \end{bmatrix}^n,
\end{aligned} \eqno{(2)}
\end{equation*}
where $x$ and $\dot{x}$ are state values, $\Delta t$ is the time interval, $n$ is the current time step, and $\bar{m}$ is the apparent mass. The matrices $\mathbf{A}_d$, $\mathbf{B}_d$ and $\mathbf{C}_d$ are state transfer matrix, state input matrix and state output matrix, respectively. The augmented model of (2) for predictive control is:

\begin{equation*}
\begin{aligned}
    \begin{bmatrix}
    \Delta \mathbf{x} \\
    y
    \end{bmatrix}^{n+1} &=
    \underbrace{
    \begin{bmatrix}
    \mathbf{A}_d & \mathbf{0}^{2 \times 1} \\
    \mathbf{C}_d \mathbf{A}_d & \mathbf{1}^{1 \times 2}
    \end{bmatrix}}_{\mathbf{A}_e}
    \underbrace{
    \begin{bmatrix}
    \Delta \mathbf{x} \\
    y
    \end{bmatrix}^n}_{\mathbf{X}^n} +
    \underbrace{
    \begin{bmatrix}
    \mathbf{B}_d \\
    \mathbf{C}_d \mathbf{B}_d
    \end{bmatrix}}_{\mathbf{B}_e} \Delta U^n, \\
    Y^n &= 
    \underbrace{
    \begin{bmatrix}
    \mathbf{0}^{1 \times 2}, 1
    \end{bmatrix}}_{\mathbf{C}_e}
    \begin{bmatrix}
    \Delta \mathbf{x} \\
    y
    \end{bmatrix}^n,
\end{aligned} \eqno{(3)}
\end{equation*}
where $\Delta \mathbf{x}$ is the increment of state $\mathbf{x}$. With $\mathbf{Y} = \left[ Y^n\ Y^{n+1} \ \hdots \ Y^{n + N_p - 1} \right]^T$, we can get a standard form of the predictive model through (3) as:

\begin{equation*}
    \mathbf{Y} = \mathbf{F} \mathbf{X}^{n} + \Phi\Delta \mathbf{U}, \eqno(4)
\end{equation*}
where:

\begin{equation*}
\begin{aligned}
    \mathbf{F} &=
    \begin{bmatrix}
    \mathbf{C}_e \mathbf{A}_e \\
    \mathbf{C}_e \mathbf{A}_e^2 \\
    \vdots \\
    \mathbf{C}_e \mathbf{A}_e^{N_p-1} \\
    \end{bmatrix},
    \\
    \Phi &=
    \begin{bmatrix}
    \mathbf{C}_e \mathbf{B}_e & 0 & \cdots & 0 \\
    \mathbf{C}_e \mathbf{A}_e \mathbf{B}_e & \mathbf{C}_e \mathbf{B}_e & \cdots & 0 \\
    \vdots \\
    \mathbf{C}_e \mathbf{A}_e^{N_p-2} \mathbf{B}_e & \mathbf{C}_e \mathbf{A}_e^{N_p-3} \mathbf{B}_e & \cdots & \mathbf{C}_e \mathbf{A}_e^{N_p-N_c-1} \mathbf{B}_e \\
    \end{bmatrix},
\end{aligned} \eqno{(5)}
\end{equation*}
In (5), $N_p$ is the prediction horizon, $N_c$ is the control horizon and $\Delta \mathbf{U} = \left[ \Delta U^n \ \Delta U^{n+1} \ \hdots \ \Delta U^{n + N_c - 1} \right]^T$. Then, we can formulate the single-axis MPC as a Quadratic Programming (QP) problem:

\begin{equation*}
\begin{aligned}
    min.& \quad J(\mathbf{E}, \Delta \mathbf{U}) =
    \underbrace{\begin{bmatrix}
    \mathbf{E} & \Delta \mathbf{U}
    \end{bmatrix}}_{\mathbf{\delta}}
    \underbrace{\begin{bmatrix}
    \mathbf{I} & \mathbf{0} \\
    \mathbf{0} & \mathbf{R}
    \end{bmatrix}}_{\mathbf{H}}
    \begin{bmatrix}
    \mathbf{E}, \\ \Delta \mathbf{U}
    \end{bmatrix}
    \\
    s.t.& \quad 
    \underbrace{\begin{bmatrix}
    \mathbf{I} & \Phi
    \end{bmatrix}}_{\mathbf{A}_{eq}}
    \begin{bmatrix}
    \mathbf{E} \\ \Delta \mathbf{U}
    \end{bmatrix} = 
    \underbrace{
    \mathbf{Y}_{ref} - \mathbf{F}\Delta \mathbf{X}^n
    }_{\mathbf{b}_{eq}}
    \\
    & \quad
    \mathbf{b}_{lb}
    \leq
    \underbrace{\begin{bmatrix}
    \mathbf{0} & \begin{bmatrix} 1 & 0 & \cdots & 0 \end{bmatrix} \\
    \begin{bmatrix} \mathbf{0} \\ \mathbf{0} \\ \vdots \\ \mathbf{0} \end{bmatrix} &
    \begin{bmatrix}
    1 & 0 & \cdots & 0 \\
    1 & 1 & \cdots & 0 \\
    \vdots \\
    1 & 1 & \cdots & 1
    \end{bmatrix} \\
    \mathbf{0} & \Phi
    \end{bmatrix}}_{\mathbf{A}}
    \begin{bmatrix}
    \mathbf{E} \\ \Delta \mathbf{U}
    \end{bmatrix} \leq
    \mathbf{b}_{ub},
\end{aligned} \eqno{(6)}
\end{equation*}
where $\delta$ denotes the optimisation variable with the tracking error $\mathbf{E}$ and the increment of control action $\Delta \mathbf{U}$. The term $\mathbf{Y}_{ref}$ denotes the desired trajectory of future $N_p$ steps, and $\mathbf{H}$ denotes the positive-defined Hessian Matrix. The matrix $\mathbf{A}_{eq}$ denotes the equality constraint matrix, $\mathbf{A}$ denotes the inequality constraint matrix, the terms $\mathbf{b}_{lb}$ and $\mathbf{b}_{ub}$ are the inequality lower and the upper bound, respectively. They can be computed as:

\begin{equation*}
    \begin{aligned}
    \mathbf{b}_{lb} &=
    \begin{bmatrix}
    \Delta \mathbf{U}_{lb} \\
    \mathbf{U}_{lb} \\
    \mathbf{Y}_{lb}
    \end{bmatrix} -
    \begin{bmatrix}
    \mathbf{0} \\
    \mathbf{U}^{n-1} \\
    \mathbf{F} \mathbf{X}^{n}
    \end{bmatrix},
    \\
    \mathbf{b}_{ub} &=
    \begin{bmatrix}
    \Delta \mathbf{U}_{ub} \\
    \mathbf{U}_{ub} \\
    \mathbf{Y}_{ub}
    \end{bmatrix} -
    \begin{bmatrix}
    \mathbf{0} \\
    \mathbf{U}^{n-1} \\
    \mathbf{F} \mathbf{X}^{n}
    \end{bmatrix}.
    \end{aligned} \eqno{(7)}
\end{equation*}

The equality constraint in (6) is used to track the reference trajectory $\mathbf{Y}_{ref}$ from the current servo step $n$ to the future servo step $n+N_p-1$, and it will provide an optimal control law $U^{n+1} = U^{n} + \Delta U^{n+1}$ with respect to the cost function by the proper setting of weight matrix $\mathbf{R} = R*\mathbf{I}$. The weight $R$ is related to the stiffness of closed-loop system, greater $R$ gives less settling time and phase delay, though it is limited by the system bandwidth. The inequality constraints are used to restrict the rate change of control action, the control action and the state output of the system. Also, notice that the terms $\mathbf{Y}_{lb} - \mathbf{F} \mathbf{X}^n$ and $\mathbf{Y}_{ub} - \mathbf{F} \mathbf{X}^n$ are the predicted distances between the bounds and the control object.

\begin{figure}
\centering
\includegraphics[width=8.65cm]{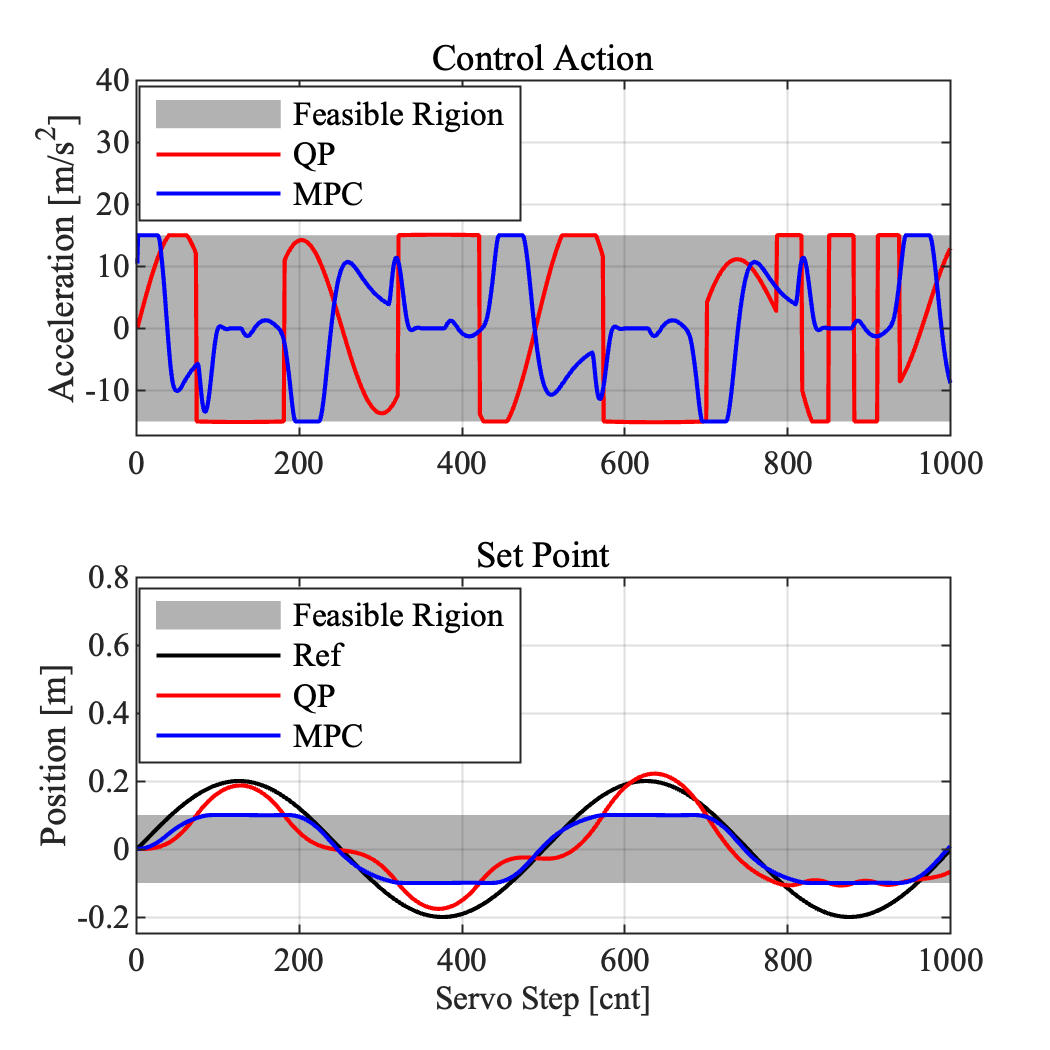}
\caption{One-step QP vs. single-axis MPC}
\label{figure:3}       
\end{figure}

Figure 3 shows the comparison between the one-step QP and the single-axis MPC by controlling an ideal one-dimensional plant. In this control system, the input bounds are [-15, 15] m/s$^2$, and the output bounds are [-0.1, 0.1] m. The system tracks a sine wave with an amplitude of 0.2 m, which is beyond the output bounds. The red lines show the performance of the QP only considering the immediate servo step. The position of the plant violates its constraints with the restriction of control action. Moreover, differentiating the control action provided by the one-step QP identifies significant jerks that actuators cannot achieve. On the other hand, the MPC controller (see blue lines) can provide feasible solutions that satisfy the output bounds with the same restriction of control action. It also shows that the single-axis MPC provides a smoother control action with small jerks.

\section{Implementation with the NSP Method}

\subsection{The NSP Method}

Since the MPC already considers inequality constraints, we use a more efficient dynamically consistent NSP whole-body controller to achieve a 500 Hz real-time servo rate instead of HQP. Sentis-Khatib proposed the dynamically consistent NSP to transform the operational control actions $\ddot{\mathbf{x}}$ to the joint torques $\mathbf{\tau}$ \cite{c22}, \cite{c20}, \cite{c23}. Kim-Sentis reported that this method costs 637 $\mu$s in mean on a dual-core 3.0 GHz CPU \cite{c25}. We can evidently save computation resources with the NSP-based method for the single-axis MPC controllers.

Consider the dynamic equations of the robot as:

$$
    \mathbf{M} \ddot{\mathbf{q}} + \mathbf{V} = \mathbf{S}^T \mathbf{\tau}, \eqno{(8)}
$$
where $\mathbf{M}$ is the mass matrix, $\ddot{\mathbf{q}}$ is the joint acceleration vector. The term $\mathbf{S}$ is the selection matrix due to floating base, $\mathbf{V}$ is the nonlinear term relating to Coriolis force, centrifugal force and gravitational force, and $\mathbf{\tau}$ is the vector of joint torques. For the dynamically consistent NSP, the torque $\mathbf{\tau}$ is:

$$
\mathbf{S}^T \mathbf{\tau} = \mathbf{J}_{foot}^T \mathbf{F}_{foot} + \mathbf{N}_{foot}^{T} \mathbf{\tau}_0, \eqno{(9)}
$$
where $\mathbf{J}_{foot}$ is the Jacobian matrix from the floating base to the foot, $\mathbf{N}_{foot}$ is the dynamically consistent null space of the matrix $\mathbf{J}_{foot}$, $\mathbf{F}_{foot}$ is the force exerted by the foot on the ground, $\mathbf{\tau}_0$ is the force of the adjacent tasks.

The mapping of the operational space acceleration and the joint space acceleration is expressed as,

$$
\ddot{\mathbf{x}}_{foot} = \mathbf{J}_{foot} \ddot{\mathbf{q}} + \dot{\mathbf{J}}_{foot} \dot{\mathbf{q}}
\eqno{(10)}
$$
where $\ddot{\mathbf{x}}_{foot}$ is the acceleration of the foot in 6 degree-of-freedoms. The foot task is usually the highest priority for the whole-body control of a humanoid robot. The NSP-based method usually considers that the support foot is virtually constrained on the ground to initiate the recursive computation. Therefore, the original research assumed that $\ddot{\mathbf{x}}_{foot}=0$, and it cancels the force $\mathbf{F}_{foot}$. However, the dynamic parameters of the real robot and the model can be different, and the computed $\mathbf{F}_{foot}$ does not guarantee $\ddot{\mathbf{x}}_{foot}=0$. Hence, we apply the mixed control to the task $\ddot{\mathbf{x}}_{foot}$ and give the operational space dynamics as:

\begin{equation*}
\begin{aligned}
\Lambda_{foot} \mathbf{U}_{foot} - \Lambda_{foot} \dot{\mathbf{J}}_{foot} \dot{\mathbf{q}} +
\Lambda_{foot} \mathbf{J}_{foot} \mathbf{M}^{-1} \mathbf{V} \\
 = \mathbf{F}_{foot} + \Lambda_{foot} \mathbf{J}_{foot} \mathbf{M}^{-1} \mathbf{N}_{foot}^{T} \mathbf{\tau}_0,
\end{aligned} \eqno{(11)}
\end{equation*}
where $\Lambda_{foot} = (\mathbf{J}_{foot} \mathbf{M}^{-1} \mathbf{J}_{foot}^T)^{-1}$ is the apparent mass matrix in the operational space and $\mathbf{U}_{foot}$ is the control action provided by the mixed control.

Equation (11) is the essential task for WBC, but it has two unknown terms, the force $\mathbf{F}_{foot}$ and the joint torques $\mathbf{\tau}$. Control actions of the task with lower priority do not interfere with the support foot task, requiring that:

$$
\Lambda_{foot} \mathbf{J}_{foot} \mathbf{M}^{-1} \mathbf{N}_{foot}^{T} \mathbf{\tau}_0 = \mathbf{0}. \eqno{(12)}
$$
Due to the existence of other tasks with lower priority, i.e., $\mathbf{\tau}_0 \neq 0$, it must satisfies that:

$$
 \Lambda_{foot} \mathbf{J}_{foot} \mathbf{M}^{-1} \mathbf{N}_{foot}^{T} = \mathbf{0}. \eqno{(13)}
$$
and, the dynamically consistent inverse of $\mathbf{J}_{foot}$ that satisfies (13) is of the form:

$$
\bar{\mathbf{J}}_{foot}^T = \Lambda_{foot} \mathbf{J}_{foot} \mathbf{M}^{-1}. \eqno{(14)}
$$

Then, the operational space dynamics of the adjacent tasks with lower priority can be expressed as:

\begin{equation*}
\begin{aligned}
\mathbf{F}_{k|k-1} &= \left( \mathbf{J}_k \mathbf{M}^{-1} \left( \mathbf{S} \mathbf{N}_{foot} \right)^T {\mathbf{J}^*_{k|k-1}}^T \right)^{-1}  \Bigg( \ddot{\mathbf{x}}_k - \dot{\mathbf{J}} \dot{\mathbf{q}} \\
& + \mathbf{J}_k \mathbf{M}^{-1} \mathbf{N}^T_{foot} \mathbf{V} + \mathbf{J}_k \mathbf{M}^{-1} \mathbf{J}^T_{foot} \Lambda^T_{foot} \dot{\mathbf{J}}_{foot}\dot{\mathbf{q}} \\
& - \mathbf{J}_k \mathbf{M}^{-1} \left( \mathbf{S} \mathbf{N}_{foot} \right)^T \Big( \mathbf{\tau}_1 + \sum_{i=2}^{k-1} \mathbf{\tau}_{i|i-1}+\mathbf{\tau}_{k|k-1} \Big) \\
& + \underbrace{\mathbf{J}_k \mathbf{M}^{-1} \mathbf{J}^T_{foot} \Lambda^T_{foot} \mathbf{U}}_{additional\ term}\Bigg),
\end{aligned} \eqno{(15)}
\end{equation*}
where ${\mathbf{J}^*_{k|k-1}}$ is termed as prioritized Jacobian for the $k^{\textup{th}}$ priority, and the subscript $k|k-1$ represents tasks with lower priority are in the null space tasks with higher priority. A detailed derivation of (15) can be seen in \cite{c21}. This paper extends (15) with an additional term since we apply the mixed control to the $\mathbf{F}_{foot}$. This additional term modifies the control action of every tasks with lower priority and ensures that the strict hierarchy of NSP will not be broken. Furthermore, the mixed control regulates the task $k$ as operational space acceleration $\ddot{\mathbf{x}}_k$ via (1).

\subsection{Implementation for ZMP}

With the application of (6), we can consider the inequality constraints with the NSP-based whole-body controller. The bounds of $\Delta \mathbf{U}$ and $\mathbf{U}$ relate to the jerk/force constraints, and the bounds of $\mathbf{Y}$ relates to the motion constraints. For the task with the highest priority, which is the support foot task, the conventional procedure assumed $\ddot{\mathbf{x}}_{foot} = \mathbf{0}$. Instead, we do not fully adopt the assumption for support foot but use the mixed control to bring the ZMP constraints under consideration. Moreover, the closed-loop control of the foot using mixed control improves the problems caused by the inconsistency of the mass parameters of the model and the actual robot and the improper execution of the joint actuators. 

However, the control actions of all directions are co-related since the apparent mass matrix $\Lambda_{foot}$ is not diagonal. Therefore, we need to apply certain approximations to fit the single-axis MPC. Thus, we express the ground reaction force (GRF) of the force $\mathbf{F}_{foot}$ in the form:

\begin{equation*}
    \mathbf{F}_{grf} = \bar{\mathbf{F}}_{grf} - \Lambda_{foot} \mathbf{U}, \eqno{(16)}
\end{equation*}
where $\bar{\mathbf{F}}_{grf} = \{\bar{F}_x, \bar{F}_y, \bar{F}_z, \bar{\tau}_x, \bar{\tau}_y, \bar{\tau}_z \}$ is the ideal GRF calculated with respect to the assumption $\ddot{\mathbf{x}}_{foot} = \mathbf{0}$. The controller $\mathbf{U}$ ensures the virtual constraint between the support foot and its contact surface by modifying the ideal GRF with the mixed control. Notice that the X-axis goes from the robot's right to left, the Z-axis from back to front, and the Y-axis from down to up in our coordinate system. Equation (16) can be substituted into the definition of friction cone and ZMP \cite{c28} to calculate the bounds for (7) as:

\begin{equation*}
\mathbf{f}_{lb} \leq \mathbf{A}' \mathbf{F}_{grf} \leq \mathbf{f}_{ub}, \eqno{(17)}
\end{equation*}
where $\mathbf{A}'$ is the friction coefficient, $\mathbf{f}_{lb}$ and $\mathbf{f}_{ub}$ are its corresponding bounds. Then, we conduct minor modifications to (17) to avoid the inverse of $\mathbf{A}'\Lambda_{foot}$ since their product is not square. Intuitively, we can use the GRF of the previous servo step by applying:

\begin{equation*}
\begin{aligned}
    \begin{bmatrix}
    -\mu \bar{F}_y \\
    0 \\
    -\mu \bar{F}_y \\
    -d_z \bar{F}_y \\
    -\mu' \bar{F}_y \\
    -d_x \bar{F}_y
    \end{bmatrix}^{n-1}  \leq
    \bar{\mathbf{F}}_{grf}^{n-1} & -\Lambda_{foot} \mathbf{U} \leq
    \begin{bmatrix}
    \mu \bar{F}_y \\
     +\infty \\
    \mu \bar{F}_y \\
    d_z \bar{F}_y \\
    \mu' \bar{F}_y \\
    d_x \bar{F}_y 
    \end{bmatrix}^{n-1},
\end{aligned} \eqno{(18)}
\end{equation*}
where the inverse of matrix $\Lambda_{foot}$ can be computed as $\mathbf{J}_{foot} \mathbf{M}^{-1} \mathbf{J}_{foot}^T$, $\mu$ and $\mu'$ are static friction coefficients, $d_x$ and $d_z$ are half lengths of foot.

\subsection{Implementation for Joint Tasks}

We can also use the proposed strategy for inequality constraints in the joint space. The formulation of the Jacobian matrix that directly maps the joint space and task space is:

\begin{equation*}
    \mathbf{J}_{jnt\{i\}} = \{ 0 \ 0\ \hdots \ \underbrace{1}_{j} \ 0 \ 0 \ \hdots \ 0 \},
    \eqno{(19)}
\end{equation*}
where $\mathbf{J}_{jnt} \in R^{u,v}$ is the Jacobian matrix of joint tasks, $u$ is the number of joints under consideration, and $v$ is the total number of joints. Setting the corresponding element of the considered joint (e.g., the elbows and the knees) $j$ as 1 directly maps their corresponding joints into the task space. Then, we can restrict the joint torques by using their apparent inertia matrix $\Lambda_{jnt}^{-1}=\mathbf{J}_{jnt} \mathbf{M}^{-1} \mathbf{J}_{jnt}^{T}$ similar to (13). Moreover, setting the bounds of $\mathbf{Y}$ in (7) can limit the movement of the joint, but it requires the prescribed set-point $\mathbf{Y}_{ref}$. We can obtain it by calculating the forward dynamics as:

$$
\ddot{\mathbf{q}}_{des} = \mathbf{M}^{-1} \mathbf{N}_{foot}^T (\mathbf{\tau} - \mathbf{V}) - \mathbf{J}_{foot}^T \Lambda_{foot} \dot{\mathbf{J}}_{foot} \dot{\mathbf{q}}. \eqno{(20)}
$$
 Equation (20) gives the desired joint accelerations $\ddot{\mathbf{q}}_{des}$, and the numerical integration can provide the desired joint position $\mathbf{q}_{des}$, which can be assigned to $\mathbf{Y}_{ref}$.

\section{Simulations and Experiment}

We conducted simulations of whole-body compliance to compare the mixed control with the NSP algorithm and the conventional HQP. We wrote the mixed control and the NSP-based whole-body controller in C++, the RBDL library \cite{c29} computes the multi-body dynamics, and qpOASES \cite{c30} solves the defined QP problems. The simulations demonstrate the advantages of the mixed control with no computation time restrictions for either the proposed algorithm or the HQP. However, considering the comparison of the two methods on the actual machine could damage our robot, in the robot experiment, we only performed the proposed algorithm to demonstrate its whole-body compliance function and time efficiency for 500 Hz real-time control.

\subsection{Simulations in Webots}

The simulation environment is the open-source robot simulator Webots. The robot model is the UBTech Walker X. In this simulation, the robot's hands can be dragged in any direction. The CoM of the whole body can also lean in the coronal and sagittal direction to provide a bigger operational space for the hands. The compliant behaviour requires a fast response to the dragging force. Without considering the above inequality constraints, the robot falls over due to the unsatisfaction of ZMP and reaching an undesired posture.  

We conducted two simulations for comparison. In the first simulation, we formulated and applied the conventional HQP for whole-body compliance. The second one utilised the proposed mixed control with NSP-base whole-body controller. These were the only difference for the experimental set-ups, and the computation time was out of consideration. In the HQP, the excessive movement of the total CoM can break the ZMP criteria. It led to the divergence of the control system and caused severe damage to the robot, see Figure 1.

\begin{figure}[b]
\centering
\includegraphics[width=8.65cm]{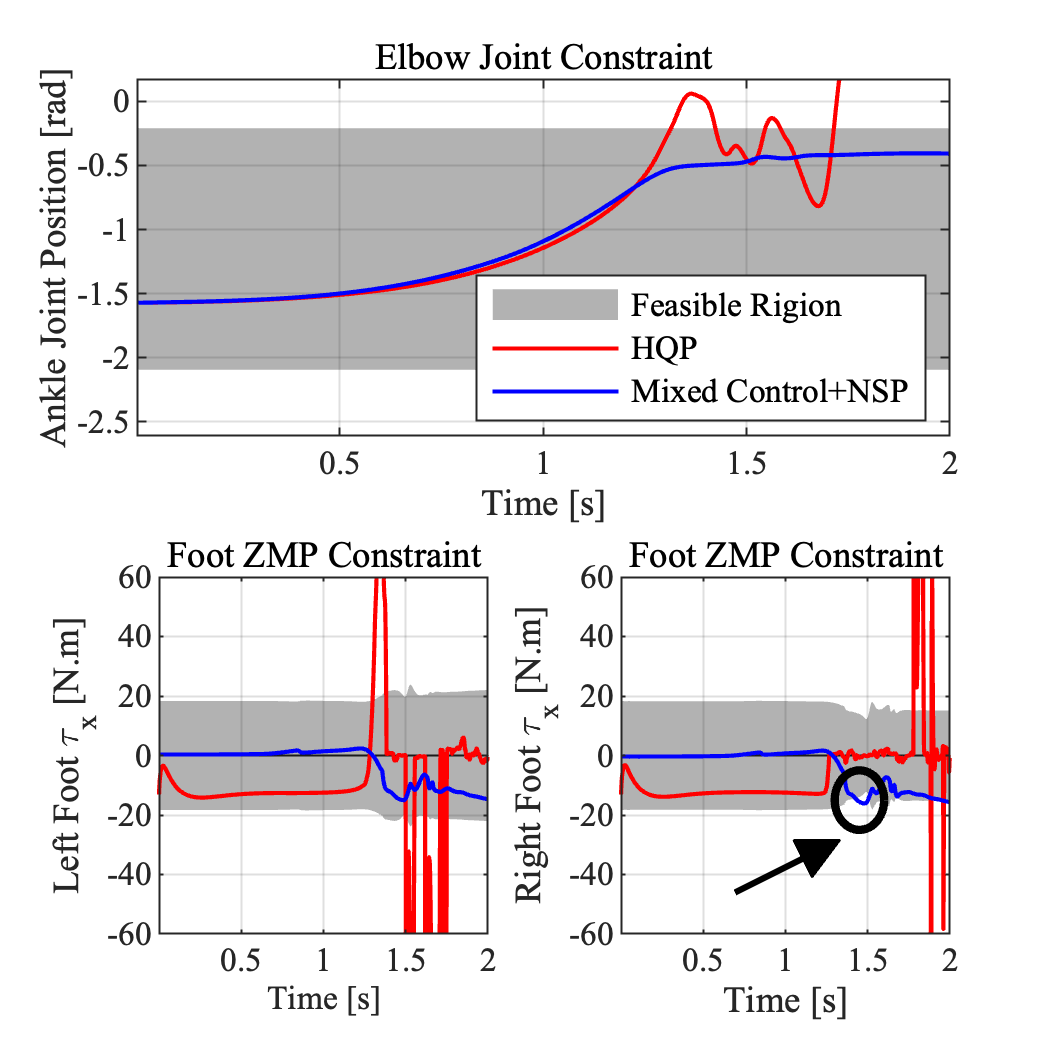}
\caption{HQP vs. the mixed control with NSP}
\label{figure:4}       
\end{figure}

\begin{figure*}
\centering
\includegraphics[width=1.0\textwidth]{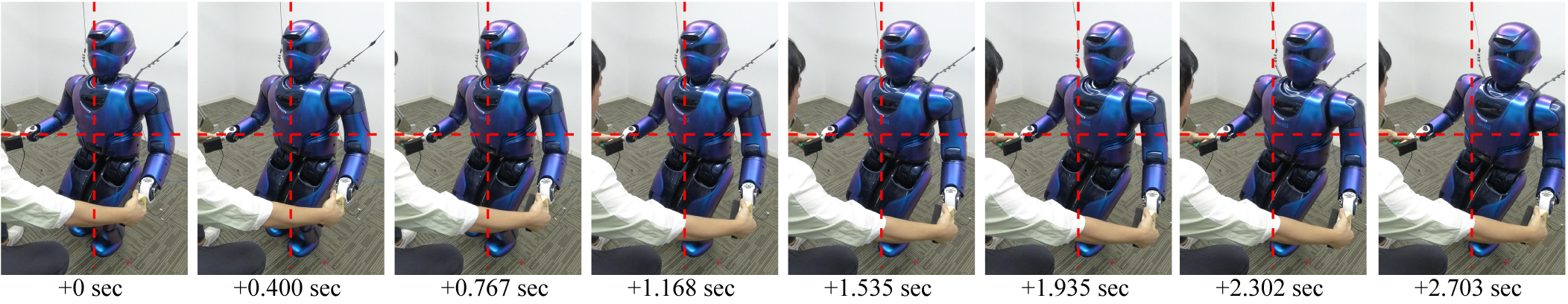}
\caption{Keyframes of the whole-body compliance experiment video Clip}
\label{figure:5}       
\end{figure*}

During the compliant behaviour, the arm is singular when the elbow joint goes to 0 rad, and an odd configuration can occur. So, the elbow joint position has to be constrained. We prioritised the joint task between the CoM task and the hand task in the two simulations. We deliberately set the joint bounds as [-2.618, -0.436] rad for the ease of demonstration. The HQP method worked when we applied the external force to make the hand move 0.3 m in 2 seconds. However, when we increased the force to make the hand move 0.4 m in 2 seconds,  the HQP method refused to satisfy the desired constraints, plotted as the red lines in Figure 4. But, the red line in the upper shows that the joint angle generally satisfied the inequality constraints by applying the mixed control and the NSP-based WBC controller.

Figure 4 also shows the ZMP constraints shaded in grey. The robot's weight is about 60 Kg, and the length of the foot width is 0.16 m. They provide the bounds [-23.5, +23.5] N$\cdot$m for the torque $\tau_{x}$ if the robot uniformly distributes its weight to both feet. As the total CoM leaning left, the ZMP of the left foot expands, but the ZMP of the right foot shrinks. The red line shows that, with one-step HQP, $\tau_{x}$ violated the ZMP limit. So when the foot flipped, the task-space torque plotted in the red line diverged at about 1.5 seconds, the abnormal behaviour of the robot stopped the simulation. With the mixed control, the trajectory of $\tau_{x}$ was within the area of ZMP, as can be seen in the plot of blue lines. It should be aware that minor violations of constraints (circled in black) can be identified in Figure 4 because the applied single-axis inertia model is still a reduced-order model. The inconsistency of models brings the small error.

\subsection{The Experiment}

Figure 5 shows the keyframes of the supplemented video, which recorded the experiment. We used (20) to obtain the position and the velocity control commands to improve the performance. The robot's wrists are equipped with 6-axis force/torque sensors to measure the external disturbance forces. The admittance controller transforms measured forces to the motion in the task space, and we compensated the mass of hands beforehand. If the robot's hand moves too far to comply with the force, a fraction of the force drives the movements of the CoM and upper-body pose to expand operational space for hands.

\begin{table}[b]
\caption{Computation Time of the Mixed Control and NSP-based Whole-Body Controller}
\label{table_example}
\begin{center}
\begin{tabular}{|c||c||c||c|}
\hline
Module & Thread & Average [{$\mu$}s] & Maximum [{$\mu$}s]\\
\hline
\multirow{2}*{Mixed Control} & Main & 535 & 867\\
\cline{2-4}
& Child &  239 & 673\\ 
\hline
NSP & Main & 440 & 790 \\
\hline
\multicolumn{2}{|c||}{Total} & 962 & 1623 \\
\hline
\end{tabular}
\end{center}
\end{table}

Table I shows the computation time of each module. As previously mentioned, we decoupled and approximated the tasks and applied the mixed control via parallel computation. In this case, the mixed control used eight single-axis MPC controllers, and the rest were PD controllers. We distributed them uniformly into the four physical cores of the Intel Core i7-8559U. Hence, this control program contains one main thread and three child threads. Moreover, we coded the NSP-based WBC into the main thread and a child thread to have similar computation times for both threads. 

The restriction of real-time control hardware is that the total computation time of the control program cannot exceed 1750 $\mu$s to achieve a 500 Hz control rate. By setting the qpQASES solver with the options {\it{setTOMPC}} and {\it{hotstart}}, the average computation time of the mixed control is 239 $\mu$s. However, if the robot's configuration changes too significant, the {\it{hotstart}} option no longer works. As a result, the maximum computation time can rise to 673 $\mu$s. In addition, data communication and scheduling between multiple threads cost extra time. Finally, the experiment shows that our maximum computation time is 1623 $\mu$s, meeting the requirement. Notice that the minimum and maximum computation times of the mixed control and NSP-based whole-body controller do not coincide. Accordingly, the total computation time is not a direct summation.

\section{Conclusions}
This paper proposed a new strategy, termed Mixed Control, for the whole-body compliance of the humanoid robot Walker X. The MPC predicts the distances between the control object and its bounds and optimises the solution several steps in advance to avoid infeasible solutions. Moreover, the jerk as the optimisation variable avoids its excessive and unrealisable magnitude, providing smooth solutions. Due to the strict hierarchical decomposition of tasks via the NSP, the application of the single-axis MPC and PD controllers in the mixed control is flexible, and the MPC controllers can be applied independently to any user-defined critical task. On the other hand, conventional PD controllers can still control other tasks. The mixed control with the NSP method runs at a 500 Hz servo rate with Intel Core i7-8559U (4-core, 1.9 GHz), which fulfils the whole-body compliance function for the humanoid robot Walker X.

Due to the limitation of computation resources, we can only apply eight MPC controllers with multi-threads for parallel computation, and the maximum computation time is about  1623 $\mu$s. It is enough for the whole-body compliance function, but the dynamic walking algorithm requires more MPC controllers. So, our future work is to improve the computational efficiency of the proposed method.






\end{document}